\documentclass[conference]{IEEEtran}
\usepackage[LGR,T1]{fontenc}
\usepackage[latin9]{inputenc}
\usepackage{array}
\usepackage{textcomp}
\usepackage{multirow}
\usepackage{amsmath}
\usepackage{amssymb}
\usepackage{graphicx}
\usepackage{rotating}

\makeatletter

\providecommand{\tabularnewline}{\\}

\IEEEoverridecommandlockouts
\usepackage{array}
\usepackage{textcomp}
\usepackage{multirow}

\usepackage{array}
\usepackage{units}
\usepackage{multirow}
\usepackage{xcolor}

\makeatother

\begin{document}
\title{Cooperation and Federation in Distributed Radar Point Cloud Processing
\thanks{Funded by the European Union. Views and opinions expressed are however
those of the author(s) only and do not necessarily reflect those of
the European Union or European Innovation Council and SMEs Executive
Agency (EISMEA). Neither the European Union nor the granting authority
can be held responsible for them. Grant Agreement No: 101099491} }
\author{\IEEEauthorblockN{S. Savazzi, V. Rampa, S. Kianoush} \IEEEauthorblockA{Consiglio Nazionale delle Ricerche (CNR)\\
 IEIIT institute, Milano\\
 Email: \{name.surname\}@ieiit.cnr.it} \and \IEEEauthorblockN{A. Minora, L. Costa} \IEEEauthorblockA{CogniMade S.r.l, https://www.cognimade.com\\
 Via C. Colombo 10/A, I-20066 Melzo, Italy\\
 Email: \{name.surname\}@cognimade.com}}
\maketitle
\begin{abstract}
The paper considers the problem of human-scale RF sensing utilizing
a network of resource-constrained MIMO radars with low range-azimuth
resolution. The radars operate in the mmWave band and obtain time-varying
3D point cloud (PC) information that is sensitive to body movements.
They also observe the same scene from different views and cooperate
while sensing the environment using a sidelink communication channel.
Conventional \emph{cooperation }setups allow the radars to mutually
exchange raw PC information to improve ego sensing. The paper proposes
a \emph{federation} mechanism where the radars exchange the parameters
of a Bayesian posterior measure of the observed PCs, rather than raw
data. The radars act as distributed parameter servers to reconstruct
a global posterior (i.e., federated posterior) using Bayesian tools.
The paper quantifies and compares the benefits of radar federation
with respect to cooperation mechanisms. Both approaches are validated
by experiments with a real-time demonstration platform. Federation
makes minimal use of the sidelink communication channel ($20\div25$
times lower bandwidth use) and is less sensitive to unresolved targets.
On the other hand, cooperation reduces the mean absolute target estimation
error of about $20\%$. 
\end{abstract}


\section{Introduction}

\label{sec:intro}

The emerging paradigm of integrated radar sensing and communication
represents a promising enabler for pervasive smart environments. These
systems target both network performance improvements \cite{poor}
and enhanced spatial discovery functions over which applications/verticals
can be optimally implemented \cite{lima}. Targeting indoor environments,
recent advances in hardware and digitalization technology enabled
the deployment of low-cost, low-power COTS (Commercial Off-The-Shelf)
radar devices of small dimension and suitable for smart home and industrial
applications \cite{computer}. In this framework, a crucial challenge
is related to the development of accurate modelling tools to optimize
the interactions between radar devices and environment as well as
among the radars themselves.

Radar technology has recently evolved to support a multitude of functions,
ranging from detection, classification, multi-subject localization
\cite{computer}, human-robot cooperation \cite{opportunistic}, human
gait, gesture, and activity recognition \cite{RADAR2,pantomime}.
Indoor classification and identification of human subjects is based
on the processing of point cloud (PC) information extracted from millimeter-wave
radars with relative low angular resolution and number of antennas
\cite{key-3,key-4}. PCs or raw data returned by the radar can be
used as input to machine learning \cite{key-3,key-2} to extract the
features of the target. Merging of PC data from multiple radar sensors
have been reported in \cite{pc_fusion} for better detection. A deep
learning approach to fuse the features of both PCs and range--Doppler
for classifying activities has been proposed in \cite{key-2}. A Bayesian
approach for fusing/grouping of radar data \cite{bayes_group} is
shown to help in angular resolution and provide added information
about the object which can be used for subsequent target/object detection.
On the other hand, the real-time exchange of radar data can be infeasible
in typical localization and tracking applications found in smart home
environments.

\begin{figure}
\centering \includegraphics[scale=0.41]{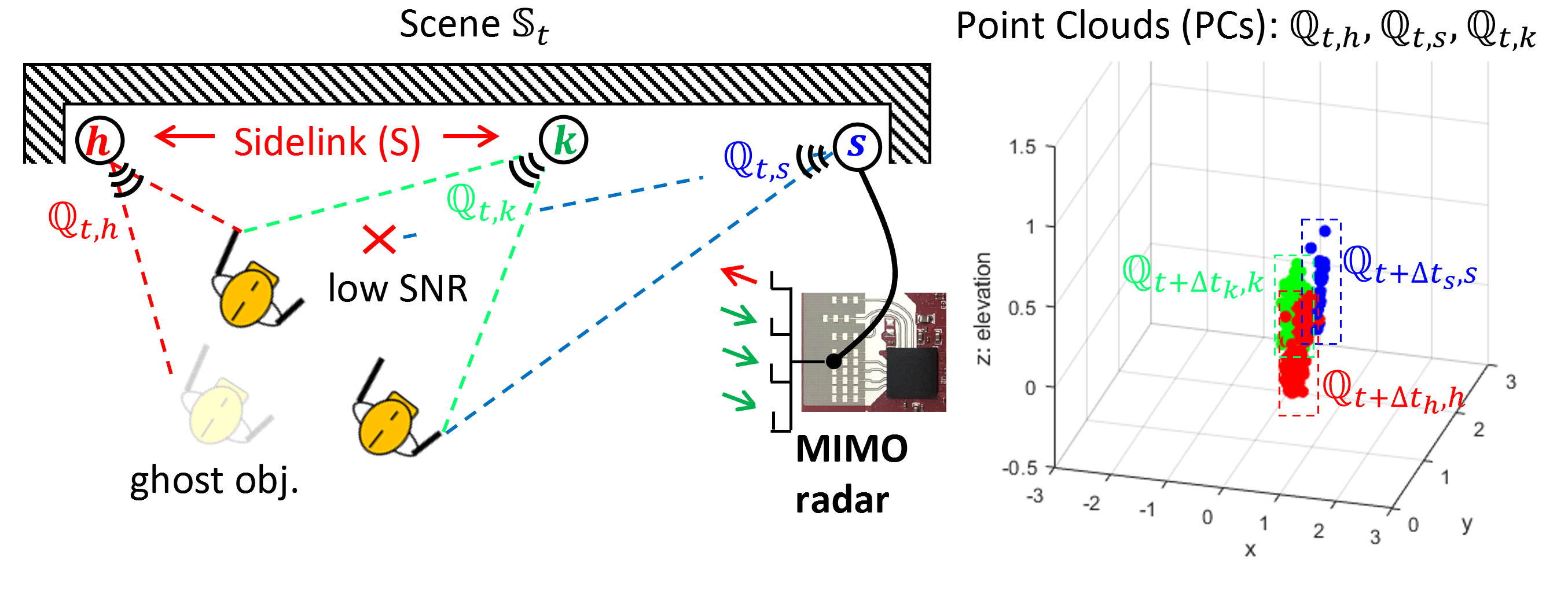} \protect\caption{\label{intro} Model of the proposed federated radar system and its
point cloud representation.}
\end{figure}

\begin{figure*}
\centering \includegraphics[scale=0.33]{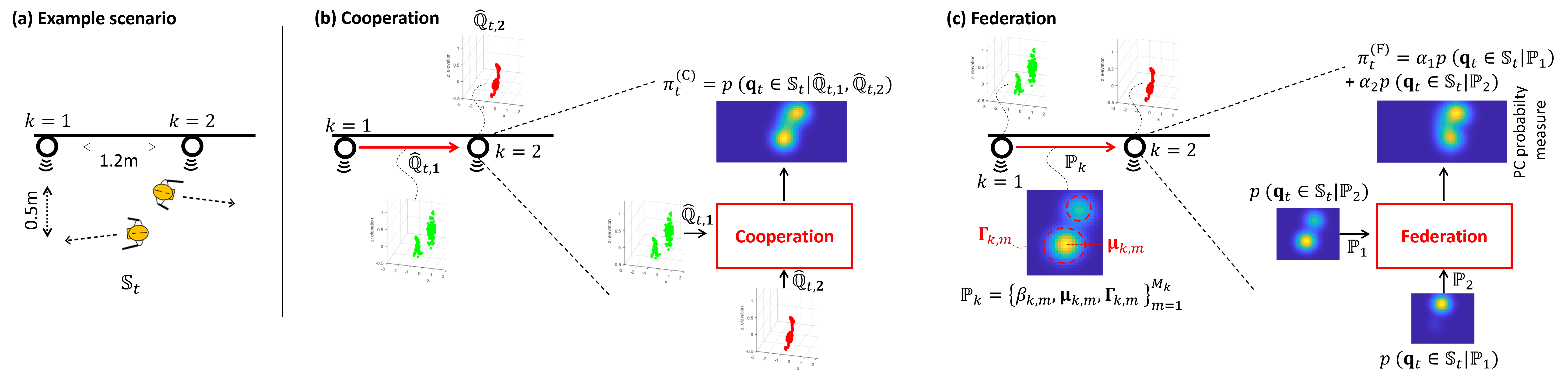} \protect\caption{\label{fedvscoop}Cooperation v. Federation PC processing architectures:
examples from real data. From left to right: (a) scene $\mathbb{S}_{t}$
consist of 2 targets co-present and moving in opposite directions
with two radars monitoring the same scene; (b) \emph{cooperation mode}:
the radars exchange pre-processed PCs $\mathbb{\widehat{Q}}_{t,k}$
and obtain a target probability measure via local fusion of the received
PCs; (c) \emph{federation mode}: the radars exchange the parameters
$\mathbb{P}_{k}$ characterizing the local posterior measure approximated
as Gaussian mixture (number of Gaussian components $M_{k}$, weights
$\beta_{k,m}$, mean $\mathbf{\boldsymbol{\mu}}_{k,m}$, and covariance
$\boldsymbol{\Gamma}_{k,m}$ of each component $m$).}
\end{figure*}

The paper investigates radar cooperation models where a sidelink communication
channel of limited bandwidth is used to coordinate the sensing task.
According to the scenario in Fig. \ref{intro}, a network of radars
is deployed to monitor the same scene $\mathbb{S}_{t}$ at time $t$.
Each radar is equipped with multiple antennas, operates in frequency
modulated continuous wave (FMCW) mode \cite{RADAR2}, and obtains
time-varying 2D/3D Point Cloud (PC) information, through local/independent
Digital Signal Processing (DSP) stages. Usage of multiple radars in
the same environment can mitigate shadowing effects of the sensed
PCs as well as improve resolution that would be obtained from ego
behavior. On the other hand, effective denoising algorithms are needed
to limit parallel/cross chirp interference sources, Signal to Noise
Ratio (SNR) degradation, blind spots, and ghost object effects. The
joint analysis of the spatio-temporal structure of the radar PCs is
here explored with the goal of improving both accuracy and resolution
of the individual radars.

\textbf{Contributions.} Unlike classical distributed radar processing
that merge/fuse raw data and/or PC information, the paper proposes
a \emph{federation model} that does not rely on data intensive fusion
of the radars raw samples on a server. The model is rooted at federation
tools \cite{mag}, and exploits sidelink communication channels of
limited bandwidths. The radars keep the raw data on the devices, thus
protecting data ownership as in as in classical federation approaches.
On the other hand, rather than exchanging the machine learning model
parameters, we propose an architecture where the radars share the
parameters of a Bayesian local posterior probability measure that
describes the (local) ``beliefs'' on the observed scene $\mathbb{S}_{t}$.
Local posteriors follow a Gaussian mixture model that is chosen for
point clouds (PC) representation. Proposed federation model is evaluated
experimentally targeting a building automation scenario. It is compared
with classical cooperation, where the radars exchange raw or pre-processed
PC info. The benefits of federation vs. cooperation are quantified
considering accuracy, resolution, and communication requirements as
well.

The paper is organized as follows. Sect. \ref{sec:mix} introduces
the system model and the PC processing tools. Cooperation and Bayesian
federation settings are analyzed in Sect. \ref{sec:Distributed-radar-settings},
while Sect. \ref{sec:Experimental-data-and} describes the developed
prototype and presents the results of the on-field experiments. Finally,
future activities are discussed in Sect. \ref{sec:conc}.

\section{Point cloud model and processing}

\label{sec:mix}As depicted in Fig. \ref{intro}, the focus is on
an indoor scenario where $K$ radio devices operate as MIMO radars
to sense and track the same time-varying scene $\mathbb{S}_{t}$ from
different points of view. The scene might be indicative of specific
body trajectories followed by one or more users (i.e., the targets)
in the monitored space, or specific actions performed by a user/subject
in a specific location(s). The observed scene is represented in terms
of (true) PCs $\mathbf{q}_{i}\in\mathbb{S}_{t}$ relative to the moving
targets in a 2D or 3D space: these points can be generally described
through an empirical probability mass function on a 2D or 3D pixel/voxel
grid. At any time $t>0,$ the scene $\mathbb{S}_{t}$ follows a stochastic
random process with probability function $p(\cdot)$ where $p(\mathbf{q}_{i}\in\mathbb{S}_{t+1}|\mathbb{S}_{t})$
is assumed to be memoryless (i.e., Markovian): therefore, the PC distribution
of the scene at time $t+1$ depends only on the scene at present time
$t$.

Each $k$-th device is equipped with a radar and measures time-varying
(3D) noisy PC information $\mathbf{\widehat{q}}_{i,k}\in\mathbb{Q}_{t,k}$.
In particular, we adopt the following PC formation model 
\begin{equation}
\mathbb{Q}_{t,k}=\left\{ \mathbf{\widehat{q}}_{i,k}\right\} =\left\{ \mathbf{q}_{i}+\mathbf{n}_{i,k}\right\} _{\mathbf{q}_{i}\in\mathbb{S}_{t}}\cup\left\{ \mathbf{o}_{j}\right\} _{\mathbf{o}_{j}\in\mathbb{O}_{t,k}},\label{eq:formation}
\end{equation}
where $\mathbf{n}_{i,k}$ is the additive noise produced by $k$-th
radar PC processing hardware, and $\mathbf{o}_{j}\in\mathbb{O}_{t,k}$
is a set of outlier points, typically caused by chirp interference,
and SNR degradation. In what follows, the PCs are obtained locally
using the Cell Averaging (CA) Constant False Alarm Rate (CA-CFAR)
method. Other approaches may be implemented as in \cite{cifars}.

The radar devices operate autonomously and independently to process
and extract local point clouds $\mathbb{Q}_{t,k}$. On the other hand,
a sidelink communication channel is used by the radars to share selected
information with the goal of improving ego knowledge. The topology
of the radar network is modelled as a directed graph $\mathcal{G}=(\mathcal{V},\mathcal{\xi})$
with the set of nodes $\mathcal{V}=\left\{ 1,...,K\right\} $ and
edges, or links, $\mathcal{\xi}=\left\{ 1,...,L\right\} $. The neighbor
set of the $k$-th device is denoted as $\mathcal{N}_{k}=\left\{ h\in\mathcal{V},h\neq k:(h,k)\in\mathcal{\xi}\right\} $,
with cardinality $\left|\mathcal{N}_{k}\right|$.

The goal of the $k$-th radar at time $t$ is to output a higher quality
reconstructed scene $\mathbb{\widehat{S}}_{t,k}$ by using side information
obtained from the radar neighborhood. In the following, we define
the ensemble of PCs in the neighbor set $\mathcal{N}_{k}$, as $\mathbb{Q}_{t,k}\cup\left\{ \mathbb{Q}_{t+\bigtriangleup t_{h},h}\right\} _{h\in\mathcal{N}_{k}},$
where $\mathbb{Q}_{t+\bigtriangleup t_{h},h}$ collects the PCs obtained
from neighbor $h\in\mathcal{N}_{k}$ and $\bigtriangleup t_{h}$ accounts
for the relative time offset w.r.t. the $k$-th radar. Compared to
the observed points $\mathbb{Q}_{t,k}$, the estimated scene $\mathbb{\widehat{S}}_{t,k}$
should be closer to the true one $\mathbb{S}_{t}$, and suitable for
further machine learning based processing, e.g. based on PointNet
methods \cite{barb,icassp}.

\section{Distributed radar models and settings}

\label{sec:Distributed-radar-settings}In the following, we compare
two distributed settings as summarized in the Fig. \ref{fedvscoop}.
The example refers to $2$ radars but can be easily generalized. For
the cooperation case (Fig. \ref{fedvscoop}(b)), the radars mutually
share a subset $\mathbb{\widehat{Q}}_{t,k}\subseteq\mathbb{Q}_{t,k}$
of the local PCs (after displacement correction and outlier removal).
Scene $\mathbb{\widehat{S}}_{t,k}$ reconstruction at radar $k$ is
obtained from the global Bayesian \emph{global posterior probability}
measure $\pi_{t}^{(\mathrm{C})}$. Considering the federation case
(Fig. \ref{fedvscoop}(c)), scene $\mathbb{\widehat{S}}_{t,k}$ reconstruction
is solved by a loose cooperation (federation) of radars that share
the parameters of a \emph{local} posterior probability\emph{ }measure,
rather than exchanging the raw points $\mathbb{\widehat{Q}}_{t,k}$.
The local posterior measures are approximated by a Gaussian mixture
function: therefore, the radars exchange the parameters $\mathbb{P}_{k}$
characterizing the mixture model (number of Gaussian components $M_{k}$,
weights $\beta_{k,m}$, mean $\mathbf{\boldsymbol{\mu}}_{k,m}$, and
covariance $\boldsymbol{\boldsymbol{\Gamma}}_{k,m}$ of each component
$m$). Scene $\mathbb{\widehat{S}}_{t,k}$ reconstruction is now obtained
from the \emph{federated posterior probability} $\pi_{t}^{(\mathrm{F})}$.
Notice that the proposed federation model has roots in Bayesian federated
learning \cite{bayes,bayes2} and decentralized setups \cite{mag}.

\subsection{Cooperation: point cloud data exchange}

\label{subsec:Cooperation}

Radars engaged in device cooperation share the PCs obtained from the
local DSP processor. Before transmission, the local points $\mathbb{Q}_{t,k}$
in (\ref{eq:formation}) are pre-processed 
\begin{equation}
\mathbb{\widehat{Q}}_{t,k}=\left\{ \mathbf{\widehat{q}}_{i,k}+\widehat{\mathbf{d}}_{i,k}\right\} _{\mathbf{\widehat{q}}_{i,k}\in\mathbb{Q}_{t,k}\setminus\mathbb{\widehat{O}}_{t,k}}\label{eq:q_hat}
\end{equation}
to correct the displacement errors $\widehat{\mathbf{d}}_{i,k}$ (i.e.,
depending on the relative position of the radar) and to remove the
outliers $\mathbb{\widehat{O}}_{t,k}$ detected in $\mathbb{Q}_{t,k}$.
The dBscan method \cite{dbscan,dbscan2} is adopted for outlier estimation:
considering the neighborhood $\mathcal{N}_{k}$, the PC ensemble $\mathbb{Q}_{t,\mathcal{N}_{k}}$
received by the $k$-th radar is 
\begin{equation}
\mathbb{Q}_{t,\mathcal{N}_{k}}=\mathbb{\widehat{Q}}_{t,k}\cup\left\{ \mathbb{\widehat{Q}}_{t+\bigtriangleup t_{h},h}\right\} _{h\in\mathcal{N}_{k}}.\label{eq:totpoints-1}
\end{equation}
Scene reconstruction is based on the \emph{global} measure $\pi_{t}^{(\mathrm{C})}$
of the posterior probability, namely the posterior obtained from the
PCs observed in the radar neighborhood: 
\begin{equation}
\pi_{t}^{(\mathrm{C})}=p(\mathbf{q}_{i}\in\mathbb{S}_{t}|\mathbb{Q}_{t,\mathcal{N}_{k}}).\label{eq:POSTERIOR}
\end{equation}
Finally, the PCs representing the true scene are here identified as
those with largest posterior probability measure: 
\begin{equation}
\mathbb{\widehat{S}}_{t,k}=\left\{ \mathbf{q}_{i}:\pi_{t}^{(\mathrm{C})}>\tau\right\} ,\label{eq:rec}
\end{equation}
with threshold $\tau$ defined at calibration time (see Sect. \ref{sec:Experimental-data-and}).
As analyzed in Sect. \ref{sec:Experimental-data-and}, the posterior
measure (\ref{eq:rec}) is used to track the positions of the targets
($\widehat{x}_{t},\widehat{y}_{t}$) in the space: these correspond
to the local maxima of the posterior, according to maximum a-posteriori
(MAP) estimation.

According to Bayesian filtering tools, the posterior $\pi_{t}^{(\mathrm{C})}$
is updated at time $t$ by an iterative approach that uses the PCs
representing the previous reconstructed scene $\widehat{\mathbb{S}}_{t-1}$,
as 
\begin{equation}
\pi_{t}^{(\mathrm{C})}\propto p(\mathbf{\widehat{q}}_{i}\in\mathbb{Q}_{t,\mathcal{N}_{k}}|\mathbb{S}_{t})\cdot p(\mathbf{q}_{i}\in\mathbb{S}_{t}|\mathbb{\widehat{S}}_{t-1,k}),\label{eq:LIK}
\end{equation}
with the conditional likelihood $p(\mathbf{\widehat{q}}_{i}\in\mathbb{Q}_{t,\mathcal{N}_{k}}|\mathbb{S}_{t})$
and the prior measure $p(\mathbf{q}_{i}\in\mathbb{S}_{t}|\mathbb{\widehat{S}}_{t-1,k})$.
Prior and conditional likelihood models suitable for tracking of human
subjects are discussed in the following sections. The interested reader
may also refer to \cite{bayesian,pointclean}. Finally, notice that
$p(\mathbf{q}_{i}\in\mathbb{S}_{t}|\mathbb{\widehat{S}}_{t-1,k})$
approximates the a-priori probability term $p(\mathbf{q}_{i}\in\mathbb{S}_{t}|\mathbb{Q}_{t-1,\mathcal{N}_{k}})=p(\mathbf{q}_{i}\in\mathbb{S}_{t}|\mathbb{S}_{t-1})\cdot\pi_{t-1}^{(\mathrm{C})}$. 

\subsection{Federation: local posterior exchange}

\label{subsec:Federation}

Rather than exchanging the PC set $\mathbb{\widehat{Q}}_{t,k}$, in
the proposed federation model the radars share the parameters of the
\emph{local posterior }probability\emph{ }measure $p(\mathbf{q}_{i}\in\mathbb{S}_{t}|\widehat{\mathbb{Q}}_{t,k})$.
The PC samples are kept on the individual radars and not transferred:
therefore, the global posterior $\pi_{t}^{(\mathrm{C})}$ in (\ref{eq:POSTERIOR})
is now approximated by a mixture of the local posteriors to obtain
$\pi_{t}^{(\mathrm{F})}$ as 
\begin{equation}
\pi_{t}^{(\mathrm{F})}=\alpha_{k}\cdot p(\mathbf{q}_{i}\in\mathbb{S}_{t}|\mathbb{\widehat{Q}}_{t,k})+\sum_{h\in\mathcal{N}_{k}}\alpha_{h}\cdot p(\mathbf{q}_{i}\in\mathbb{S}_{t}|\mathbb{\widehat{Q}}_{t,h}),\label{eq:fed}
\end{equation}
where $\alpha_{k}+\sum_{h\in\mathcal{N}_{k}}\alpha_{h}=1$ and $\alpha_{k}=\frac{Q_{k}}{Q_{k}+\mathop{\sum_{h}Q_{h}}}$.
$Q_{k}$ is the number of points measured by radar $k$. The \textit{federated
posterior} $\pi_{t}^{(\mathrm{F})}$ can be evaluated either centrally
by a server collecting the local posterior parameters, namely the
Parameter Server, (PS) or distributedly. Scene reconstruction is represented
similarly as in (\ref{eq:rec}), now with $\mathbb{\widehat{S}}_{t,k}=\left\{ \mathbf{q}_{i}:\pi_{t}^{(\mathrm{F})}>\tau\right\} $.
Notice that, the estimated scene $\mathbb{\widehat{S}}_{t,k}$ from
the federated posterior is used to update the \emph{local} posterior
on the next step $t+1$, thus augmenting the $k$-th radar knowledge
about the monitored space. Likewise equation (\ref{eq:LIK}), the
local posterior can be also evaluated iteratively as $p(\mathbf{q}_{i}\in\mathbb{S}_{t}|\widehat{\mathbb{Q}}_{t,k})\propto p(\mathbf{q}_{i}\in\mathbb{S}_{t}|\mathbb{\widehat{S}}_{t-1,k})\cdot p(\mathbf{\widehat{q}}_{i}\in\widehat{\mathbb{Q}}_{t,k}|\mathbb{S}_{t})$
using the prior $p(\mathbf{q}_{i}\in\mathbb{S}_{t}|\mathbb{\widehat{S}}_{t-1,k})$
and the (local) conditional likelihood $p(\mathbf{\widehat{q}}_{i}\in\widehat{\mathbb{Q}}_{t,k}|\mathbb{S}_{t})$,
respectively.

Cooperation (\ref{eq:POSTERIOR}) versus federation (\ref{eq:fed})
tradeoffs are discussed in Sect. \ref{sec:Experimental-data-and}
targeting the problem of multi body tracking. In what follows, we
adopt the multivariate Gaussian mixture distributions to represent
both the local, the global posterior probability measures, and the
priors.

\begin{figure}[t]
\centering \includegraphics[scale=0.37]{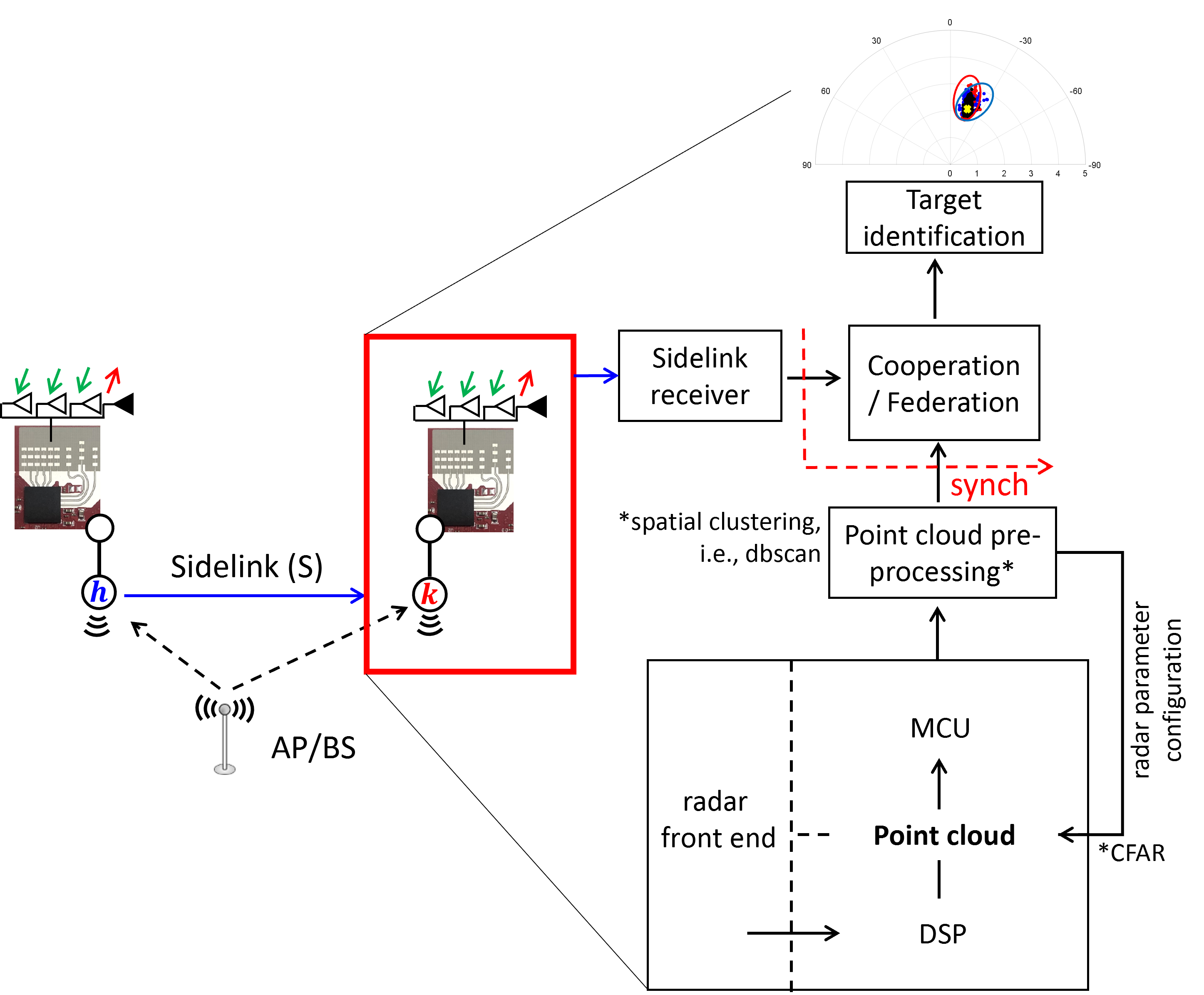} \protect\caption{\label{architecture}Proposed architecture and system model. Radar
$k$ monitoring the sidelink (S) and receiving information from radar
$h$. The radar MIMO HW supports range-azimuth and elevation tracking
and it is equipped with a DSP to extract time-varying 2D/3D point
cloud information. The radar CFAR configuration parameters can be
controlled via UART communication with the MCU. }
\end{figure}

\begin{figure}[t]
\centering \includegraphics[scale=0.22]{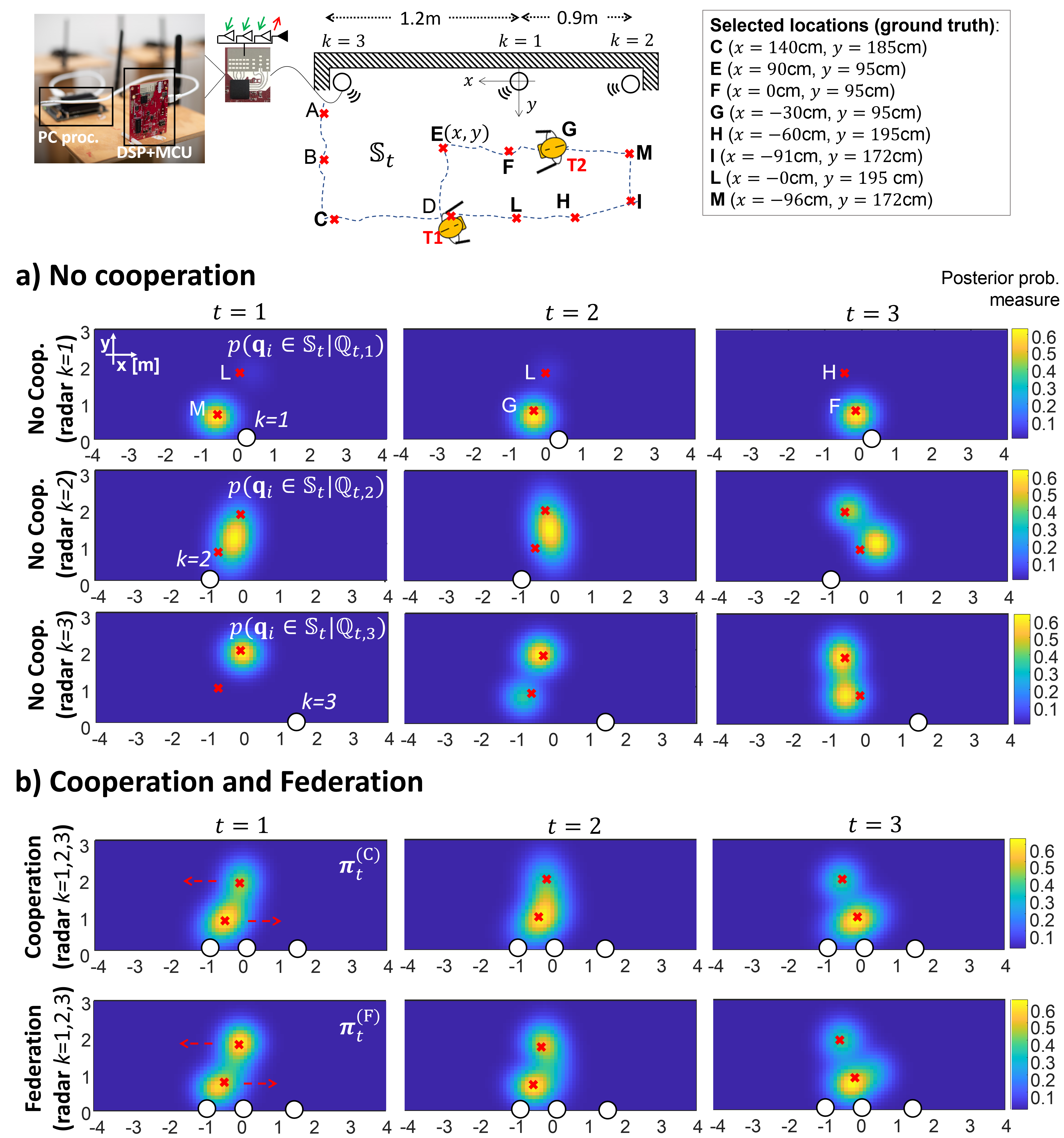} \protect\caption{\label{scene}Scene example featuring 2 subjects moving from position
L to H and from M to F. Posteriors observed over three consecutive
time samples ($t=1,2,3$): a) no-cooperation case with $k=1,2,3$;
and b) federation and cooperation cases for $K=3$ radars with deployment
described above. }
\end{figure}

\begin{figure}
\centering \includegraphics[scale=0.64]{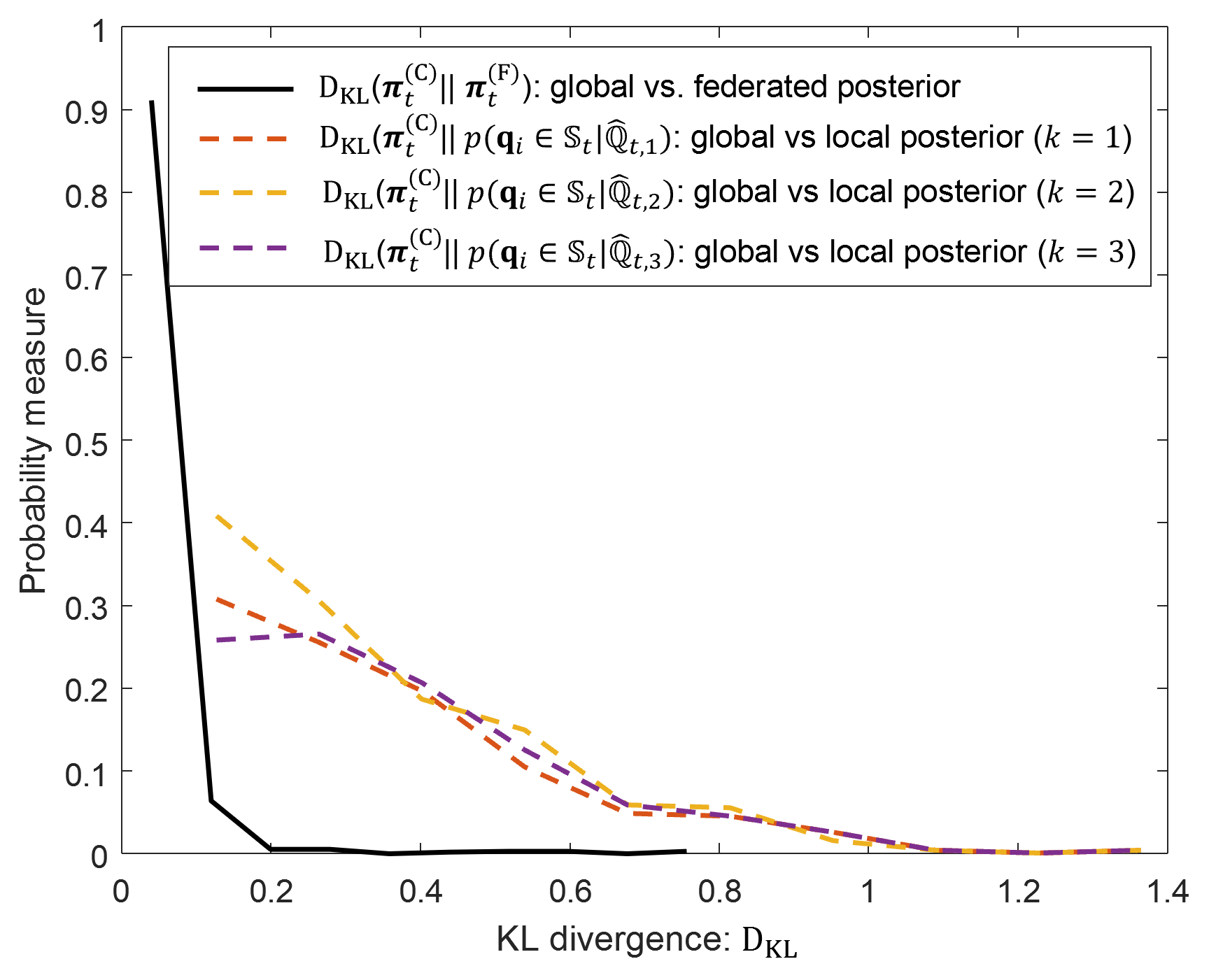} \protect\caption{\label{kull} Kullback-Leibler (KL) $\mathrm{D_{KL}}[\cdot]$ divergence
sample probability functions $\mathrm{Pr}[\mathrm{D_{KL}}]$: solid
line compares global and federated posteriors $(\mathrm{D_{KL}}[\pi_{t}^{(\mathrm{C})}\Vert\pi_{t}^{(\mathrm{F})}]$),
dashed lines compare global and local posteriors for the three deployed
radars. Probability measure is obtained by collecting $900$ consecutive
sample divergences ($9$ sec. with $10$ms per localization update)
corresponding to $2$ subjects moving from position L to H and from
M to F (same scenario as in Figure \ref{scene}).}
\end{figure}

\subsection{Gaussian mixture approximations}

Local/global posteriors follow a multivariate Gaussian mixture model.
The local posterior is therefore approximated as 
\begin{equation}
p(\mathbf{q}_{i}\in\mathbb{S}_{t}|\widehat{\mathbb{Q}}_{t,k})\simeq\sum_{m=1}^{M_{k}}\beta_{k,m}\cdot\mathcal{N}(\mathbf{\mu}_{k,m},\mathbf{\Sigma}_{k,m}),\label{eq:multiva}
\end{equation}
with parameters $\mathbb{P}_{k}=\left\{ \beta_{k,m},\mathbf{\boldsymbol{\mu}}_{k,m},\mathbf{\boldsymbol{\Sigma}}_{k,m}\right\} _{m=1}^{M_{k}}$
for each $k$-th radar. $M_{k}$ is the number of components observed
by radar $k$, while ($\mathbf{\boldsymbol{\mu}}_{k,m},\mathbf{\boldsymbol{\Sigma}}_{k,m}$)
are the mean and covariance of the $m$-th component\footnote{for 3D PCs, $\mathbf{\boldsymbol{\mu}}_{k,m}$ has dimension $3\times1$
and covariance $\mathbf{\boldsymbol{\Sigma}}_{k,m}$ is $3\times3$.} and $\sum_{m=1}^{M_{k}}\beta_{k,m}=1$. Number of components $M_{k}$
depends on the post-processed PCs $\widehat{\mathbb{Q}}_{t,k}$ after
outlier removal \cite{dbscan2} and will be assessed in the following
experimental analysis. The federation model (\ref{eq:fed}) requires
the radars share the parameters $\mathbb{P}_{k}$ of the Gaussian
mixture (\ref{eq:multiva}). Federated posterior measure $\pi_{t}^{(\mathrm{F})}$
is thus obtained by substituting (\ref{eq:multiva}) in (\ref{eq:fed}).
For the cooperation model, the global posterior $\pi_{t}^{(\mathrm{C})}$
is 
\begin{equation}
\pi_{t}^{(\mathrm{C})}\negthinspace=\negthinspace p(\mathbf{q}_{i}\in\mathbb{S}_{t}|\mathbb{Q}_{t,\mathcal{N}_{k}})\negthinspace\simeq\negthinspace\sum_{m=1}^{N}\beta_{\mathrm{C},m}\cdot\mathcal{N}(\mathbf{\mu}_{\mathrm{C},m},\mathbf{\Sigma}_{\mathrm{C},m}),\label{eq:GLOB}
\end{equation}
with $\sum_{m=1}^{N}\beta_{C,m}=1$, and typically $N\gg M_{k}$.
Global and federated posteriors are compared in the following section
in terms of the Kullback-Leibler (KL) divergence.

As far as the prior $p(\mathbf{q}_{i}\in\mathbb{S}_{t}|\mathbb{\widehat{S}}_{t-1,k})$
is concerned, we assume that the objects/targets in the scene consist
of smooth and well separated patches moving at maximum speed $v$.
For $S$ points in the scene $\mathbb{\widehat{S}}_{t-1,k}$, the
prior measure 
\begin{equation}
p(\mathbf{q}_{i}\in\mathbb{S}_{t}|\mathbb{\widehat{S}}_{t-1,k})\simeq\frac{1}{S}\sum_{\mathbf{q}_{i}\in\mathbb{\widehat{S}}_{t-1,k}}\mathcal{N}\left[\mathbf{q}_{i},\sigma^{2}(v)\times\mathbf{I}\right]
\end{equation}
is obtained as the superposition of Gaussian distributions each modelling
the motion of individual particle $\mathbf{q}_{i}$ with $\sigma^{2}(v)$
obtained from a random walk model \cite{random}. Other priors can
be also exploited to quantify the smoothness and sharpness of the
targets as proposed in \cite{bayesian,ml} for imaging applications.

\section{Experimental data and validation}

\label{sec:Experimental-data-and}This section discusses cooperation
and federation models based on a prototype implementation and experimental
data. Proposed architecture is summarized in Fig. \ref{architecture}:
in the example, the radar $k$ is monitoring the sidelink (S) and
receiving information from radar $h$. The radars work in the $77$
GHz band and support MIMO-FMCW radio featuring an array of $3$ TX
and $4$ uniformly spaced RX antennas. Each radar has an azimuth Field
Of View (FOV) of -/+ $60$° angle and range resolution of $25$° and
$4.2$cm, 3.9 GHz band (sweep time $36$$\mu$s). It integrates a
C674x DSP and an ARM R4F MCU with TX output power set to $12$ dBm.
PC samples are obtained from the radar DSP using the CA-CFAR method
configured with range and Doppler detection thresholds of $15$dB.
Then, PCs are further processed using a ARM-Cortex-A57 SoC (Jetson
Nano device model). Notice that the radar CFAR configuration parameters
can be controlled via UART communication with the MCU.

Three wall-mounted radars are deployed as detailed in Fig. \ref{scene}
(positions are relative to radar $k=1$). The scene $\mathbb{S}_{t}$,
detailed in Fig. \ref{scene} (on top), consists of $2$ co-present
human subjects moving inside an indoor lab environment. Trajectories
follow $6$ landmark points (A-M), with corresponding relative ground-truth
locations detailed in the same figure. Radars exchange either raw
PCs (cooperation) or local model parameters $\mathbb{P}_{k}$ (federation)
via WiFi links while synchronization is implemented using NTP (Network
Time Protocol). Impact of inaccurate clock source on performance \cite{ntp}
is here assessed experimentally: however, a dedicated analysis is
out of scope.

\subsection{Global, federated and local posterior analysis}

Figure \ref{scene} compares the local posterior distributions $p(\mathbf{q}_{i}\in\mathbb{S}_{t}|\mathbb{\widehat{Q}}_{t,k})$
for three (isolated) radars $k=1,2,3$ (Fig. \ref{scene}.a) with
the global $\pi_{t}^{(\mathrm{C})}$ and federated posterior $\pi_{t}^{(\mathrm{F})}$
(Fig. \ref{scene}.b). Posteriors are observed at $3$ consecutive
time instants $t=1,2,3$. The radar $k=1$ correctly identifies the
target T2 following the path M$\rightarrow$G$\rightarrow$F, but
it is less effective for T1, moving in the opposite direction, from
L$\rightarrow$H, due to low SNR. On the other hand, the radar $k=2$
identifies T1, while T2 is now shadowed. Cooperation and federation
cases refer to a scenario where the radar $k=1$ obtains side information
from neighbors $k=2,3$: in both cases, the radar is able to effectively
resolve both targets.

Figure \ref{kull} now assess in more detail the capability of the
federated posterior $\pi_{t}^{(\mathrm{F})}$ in (\ref{eq:fed}) to
effectively reproduce the global posterior $\pi_{t}^{(\mathrm{C})}$
as obtained by fusion of the PCs on the radar. We thus analyze and
compare the Kullback-Leibler (KL) distance $\mathrm{D_{KL}}$ \cite{kull}
between the global, the federated and the local posterior measures.
Notice that the global posterior in (\ref{eq:POSTERIOR}) approximates
the posterior that would be obtained by letting the radar exchange
the (full) raw 3D radar data. However, exchanging radar data over
sidelinks is not considered in this paper as infeasible due to bandwidth
constraints introduced by the real-time data sharing process (Fig.
\ref{architecture}). Considering that the posterior is computed on
each localization update, we collect consecutive KL distances corresponding
to $2$ subjects moving from position L to H and from M to F (see
Figure \ref{scene}) and obtain a sample probability $\mathrm{Pr}[\mathrm{D_{KL}}]$.
Solid line compares global and federated posteriors $(\mathrm{D_{KL}}[\pi_{t}^{(\mathrm{C})}\Vert\pi_{t}^{(\mathrm{F})}]$),
while dashed lines compare global and local posteriors ($\mathrm{D_{KL}}[\pi_{t}^{(\mathrm{C})}\Vert p(\mathbf{q}_{i}\in\mathbb{S}_{t}|\mathbb{\widehat{Q}}_{t,k})],$
$k=1,2,3$) for the three radars. Proposed federation setup confirms
the property of being a loose form of cooperation that makes the federated
posterior closer to the global one, compared with the local measures,
but with no exchange of the raw PCs. 
\begin{figure}
\centering \includegraphics[scale=0.62]{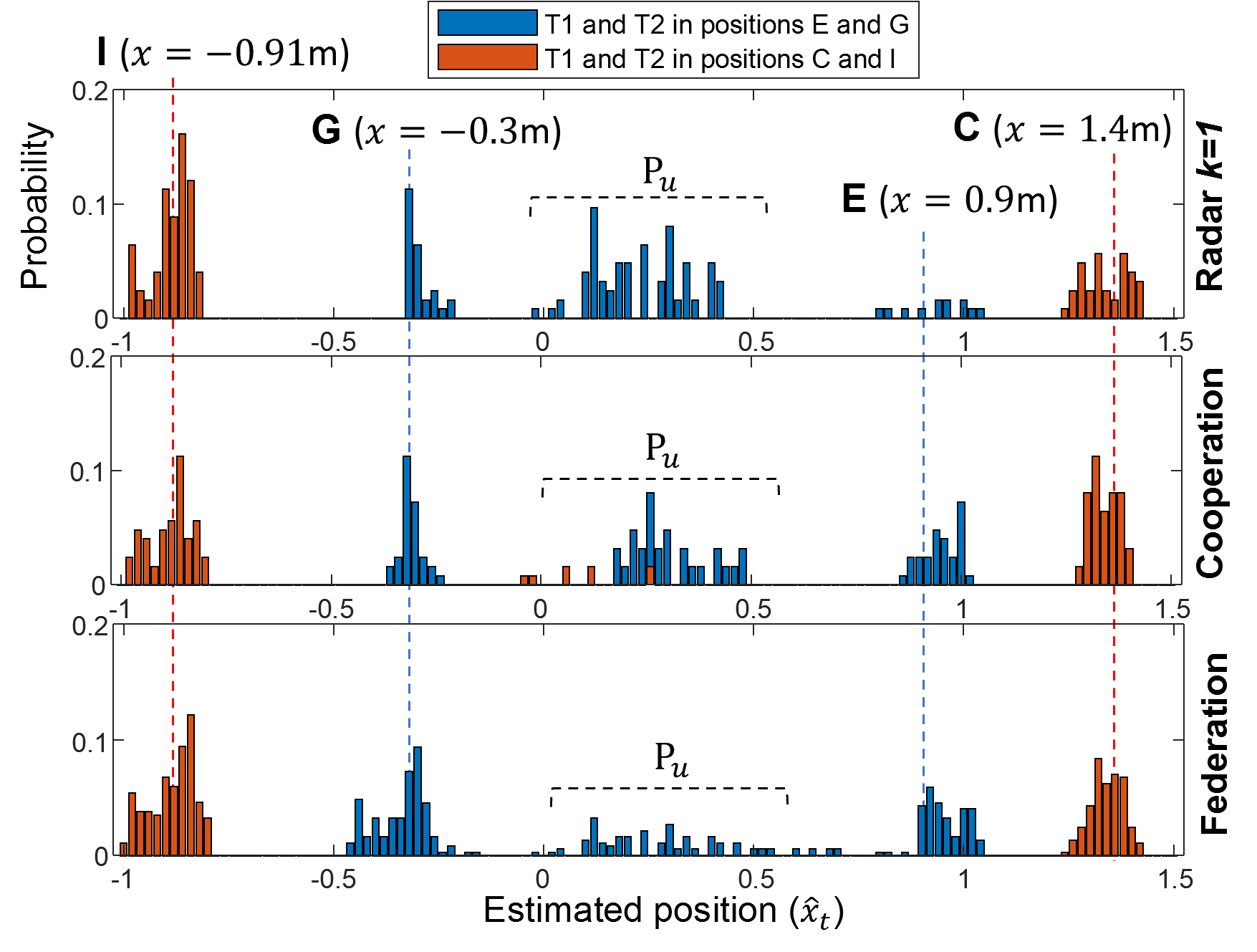} \protect\caption{\label{HIST} Probability distribution of estimated positions $\widehat{x}_{t}$
for targets T1, T2 located in C, I (red bars) moving to G, E (blue
bars). $\mathrm{P}_{u}$ is the probability of unresolved targets.
Cooperation, federation and isolated ($k=1$) cases are compared.}
\end{figure}

\begin{table}[tp]
\protect\caption{\label{table_complex_acc-1}MAE $\sigma(x),\sigma(y)$ {[}m{]}, probability
of unresolved targets ($\mathrm{P}_{u}$) and sidelink overhead $B$
{[}MB{]} considering T1 and T2 in positions (C, I) and (G, E).}
\vspace{0.3cm}

\begin{centering}
\begin{tabular}{l|l|l|l|}
\begin{turn}{90}
\end{turn} & $k=1$  & \textbf{Cooperation}  & \textbf{Federation}\tabularnewline
\hline 
\multirow{4}{*}{\begin{turn}{90}
$\sigma$ (acc.) 
\end{turn}} & I:$\underset{\sigma(x)}{0.08},\underset{\sigma(y)}{0.15}$  & I:$\underset{\sigma(x)}{0.08},\underset{\sigma(y)}{0.13}$  & I:$\underset{\sigma(x)}{0.07},\underset{\sigma(y)}{0.15}$\tabularnewline
 & C:$0.09,0.16$  & C: $0.09$$,0.07$  & C:$0.08,0.09$\tabularnewline
\cline{2-4} \cline{3-4} \cline{4-4} 
 & G:$0.23,0.14$  & G:$0.07,0.09$  & G:$0.11,0.09$\tabularnewline
 & E:$0.12,0.15$  & E:$0.08,0.08$  & E:$0.1,0.12$\tabularnewline
\hline 
\multirow{2}{*}{\begin{turn}{90}
$\mathrm{P}_{u}$ 
\end{turn}} & (C, I):$10\%$  & (C, I):$1\%$  & (C, I):$1\%$\tabularnewline
\cline{2-4} \cline{3-4} \cline{4-4} 
 & (G, E):$59.3\%$  & (G, E):$40.9\%$  & (G, E):$19.8\%$\tabularnewline
\hline 
\multirow{4}{*}{\begin{turn}{90}
$B$ {[}MB{]} 
\end{turn}} & (C, I):$0.0$  & (C, I):$12$ Mbit/s  & (C, I):$0.28$ Mbit/s\tabularnewline
 &  & $62.500$ PCs/s  & $4.400$ $\mathbb{P}_{k}$/s\tabularnewline
\cline{2-4} \cline{3-4} \cline{4-4} 
 & (G, E):$0.0$  & (G, E):$15.6$ MB/s  & (G, E):$0.28$ Mbit/s\tabularnewline
 &  & $81.500$ PCs/s  & $4.400$ $\mathbb{P}_{k}$/s\tabularnewline
\hline 
\end{tabular}
\par\end{centering}
\medskip{}
 \vspace{-0.6cm}
\end{table}

\subsection{MAE, unresolved targets and communication overhead}

Fig. \ref{HIST} and Tab. \ref{table_complex_acc-1} provide a quantitative
evaluation of the localization accuracy and communication overhead.
T1 and T2 are now located in positions C,I and move towards the corresponding
landmarks E,G, by following the trajectory C$\rightarrow$D$\rightarrow$E
and I$\rightarrow$M$\rightarrow$G, respectively. The estimated positions
of the subjects ($\widehat{x}_{t},\widehat{y}_{t}$) in the space
correspond to the local maxima of the posteriors ($\pi_{t}^{(\mathrm{C})}$,$\pi_{t}^{(\mathrm{F})}$),
according to MAP estimation. Notice that before obtaining the local
maxima, we apply a thresholding (\ref{eq:rec}) of the posteriors,
with $\tau=0.45$ to limit the search time of the local maxima to
the PCs that belong to the reconstructed scene $\mathbb{\widehat{S}}_{t,k}$.
Fig. \ref{HIST} shows the probability distribution of estimated positions
$\widehat{x}_{t}$ obtained by radar $k=1$ alone, via cooperation
and federation, respectively. Low azimuth resolution of the radar
$k=1$ causes targets in positions G,E be unresolved with large probability
($\mathrm{P}_{u}$). Cooperation limits these impairments, while federation
shows superior performance being less sensitive to the number of observed
PCs. For the same scene, Tab. \ref{table_complex_acc-1} shows the
localization mean absolute error (MAE), namely $\sigma(x)=\mathbb{E}_{t}\left[\left|\widehat{x}_{t}-x\right|\right]$,
$\sigma(y)=\mathbb{E}_{t}\left[\left|\widehat{y}_{t}-y\right|\right]$,
the probability of unresolved targets $\mathrm{P}_{u}$, and the communication
overhead $B$ {[}Mbit/s{]}, quantified here as the sum of bits exchanged
over the sidelink per second (localization updates are issued on every
$10$ ms). The number of exchanged PCs (cooperation) and the mixture
model parameters $\mathbb{P}_{k}$ (federation) per second are also
reported.

Cooperation generally improves the average accuracy ($\sigma(x)=0.09$m)
compared with federation ($\sigma(x)=0.11$m). On the other hand,
federation is more effective in minimizing unresolved targets $(\mathrm{P}_{u}$)
as less sensitive to outliers: this effect can be appreciated for
co-located subjects G-E. Federation also keeps the required communication
overhead about $20\div25$ times lower than cooperation. In cooperation
mode, the number of PCs exchanged by each radar depend on the subject
position. Subjects in positions C and I make each radar send $625$
PCs per localization update on average ($62.500$ PC/s), or $12$
Mbit/s, considering that 3D PCs are $64$-bit quantized and formatted
in JSON \cite{opportunistic}. Subjects in positions G and E are represented
on average by $815$ PCs per update with required sidelink throughput
of $15.6$ MB/s. In federation mode, each radar exchanges (on average)
$44$ local posterior parameters ($\mathbb{P}_{k}$) per localization
update, namely $281.6$ Kbit/s. Parameters exchanged contain: the
points $Q_{k}$ measured by radar $k$ and used to represent the federated
posterior as in (\ref{eq:fed}), $M_{k}=3$ of components (on average),
and, for each component, $14$ parameters namely, $\beta_{k,m}$,
$\mathbf{\boldsymbol{\mu}}_{k,m}$ (3 terms), $\mathbf{\boldsymbol{\Sigma}}_{k,m}$
(9 terms) and the number of represented PCs $Q_{k,m}$, being $\sum_{m=1}^{M_{k}}Q_{k,m}=Q_{k}$.

\section{Conclusion and future activities}

\label{sec:conc} The paper proposed and compared cooperation and
federation models suitable for distributed radar point cloud processing.
The considered radars are characterized by limited range-angular resolution
when operating in isolation; however, they exchange side information
using a communication channel of limited bandwidth. The paper discusses
a prototype and related experiments that quantify the benefits of
federation vs. cooperation targeting the localization and tracking
of two people in an indoor environment. Federation and cooperation
benefits are quantified in terms of latency/sidelink bandwidth use,
resolution, probability of unresolved targets and mean absolute localization
error performance.

Cooperation generally provides improved average mean absolute localization
error ($0.09$m) than federation ($0.11$m). Federation forces the
radar to exchange model parameters of observed PCs probability (local
posterior measure) and it is thus more effective in minimizing unresolved
targets and outliers. Federation is a promising solution for scaling
up the system to support the deployment of more radars as it keeps
the required communication overhead about $20\div25$ times lower
than cooperation. Required sidelink throughput is also independent
from the number of observed PCs. Future activities will consider the
impact of synchronization, radar network topologies and heterogeneity
in terms of angular resolution and field of view.

\end{document}